%% file: root.tex
\documentclass[letterpaper, 10pt, conference]{ieeeconf}  % Comment this line out if you need a4paper

\IEEEoverridecommandlockouts    % This command is only needed if

\usepackage{amsmath} % assumes amsmath package installed
\usepackage{bm}
\usepackage{amssymb}  % assumes amsmath package installed
\usepackage{babel}
\usepackage[style=english]{csquotes}
\usepackage{svg}
\usepackage{makecell}
\usepackage{colortbl}
\usepackage{xcolor}
\usepackage{multirow}
\usepackage{booktabs}
\usepackage{tabularx}
\usepackage{tabulary}
\let\labelindent\relax % enumitem error fix: \labelindent already defined.
\usepackage{enumitem}
\usepackage{orcidlink}
\usepackage{graphicx}
\usepackage{subcaption}

\title{\LARGE \bf
Space, Time, and Interaction: A Taxonomy of Corner Cases in Trajectory Datasets for Automated Driving
}

\author{Kevin Rösch\orcidlink{0000-0002-6841-8484}$^{1,*}$,
        Florian Heidecker\orcidlink{0000-0003-2895-0254}$^{2,*}$,
        Julian Truetsch\orcidlink{0000-0001-5824-2617}$^{1,*}$,
        Kamil Kowol\orcidlink{0000-0001-6951-7081}$^{3,*}$,\\
        Clemens Schicktanz\orcidlink{0000-0002-3234-2086}$^{4,*}$,
        Maarten Bieshaar\orcidlink{0000-0002-6471-6062}$^{5,*}$,
        Bernhard Sick\orcidlink{0000-0001-9467-656X}$^{2}$,
        Christoph Stiller\orcidlink{0000-0003-4165-2075}$^{6}$
    \thanks{* Authors contributed equally.}
    \thanks{$^{1}$Kevin Rösch and Julian Truetsch are with Mobile Perception Systems, FZI Research Center for Information Technology, Schönfeldstraße 8, 76131 Karlsruhe, Germany, {\tt\small \{kevin.roesch, truetsch\}@fzi.de}}%
    \thanks{$^{2}$Florian Heidecker and Bernhard Sick are with Intelligent Embedded Systems, University of Kassel, Wilhelmshöher Allee 73, 34121 Kassel, Germany, {\tt\small \{florian.heidecker, bsick\}@uni-kassel.de}}%
    \thanks{$^{3}$Kamil Kowol is with School of Mathematics and Natural Sciences, University of Wuppertal, Lise-Meitner-Straße 27-31, Wuppertal, Germany,  {\tt\small kowol@uni-wuppertal.de}}% <-this % stops a space
    \thanks{$^{4}$ Clemens Schicktanz is with German Aerospace Center (DLR), Institute of Transportation Systems, Rutherfordstr. 2, 12489 Berlin, Germany, {\tt\small clemens.schicktanz@dlr.de}}%
    \thanks{$^{5}$ Maarten Bieshaar is with Robert Bosch GmbH, 31132 Hildesheim, Germany, {\tt\small maarten.bieshaar@de.bosch.com}}
    \thanks{$^{6}$ Christoph Stiller is with Institute  of  Measurement  and  Control  Systems, Karlsruhe Institute of Technology, Engler-Bunte-Ring 21, 76131 Karlsruhe, Germany, {\tt\small stiller@kit.edu}}%
}

\definecolor{light blue}{HTML}{77AADD}
\definecolor{light cyan}{RGB}{10,171,231}
\definecolor{mint}{HTML}{44BB99}
\definecolor{pear}{HTML}{BBCC33}
\definecolor{olive}{HTML}{AAAA00}
\definecolor{light yellow}{HTML}{EEDD88}
\definecolor{light red}{HTML}{EE8866}
\definecolor{pink}{HTML}{FFAABB}
\definecolor{pale grey}{HTML}{DDDDDD}

\definecolor{perception}{RGB}{204,204,41}
\definecolor{decision}{RGB}{163,204,41}
\definecolor{goal}{RGB}{95,163,0}
\definecolor{execution}{RGB}{41,204,41}
\definecolor{body}{RGB}{41,204,163}
\definecolor{attention}{RGB}{41,190,204}
\definecolor{allocation}{RGB}{155,201,204}
\definecolor{knowledge}{RGB}{41,150,204}
\definecolor{recording}{RGB}{204,122,149}
\definecolor{environmentinfo}{RGB}{177,122,204}
\definecolor{environment}{RGB}{242,101,73}

\begin{document}
\newcommand{\fh}[1]{{\color{olive} #1}}
\newcommand{\jt}[1]{{\color{blue} #1}}
\newcommand{\kk}[1]{{\color{green} #1}}
\newcommand{\mb}[1]{{\color{orange} #1}}
\newcommand{\kr}[1]{{\color{magenta} #1}}
\newcommand{\cs}[1]{{\color{teal} #1}}
\newcommand{\js}[1]{{\color{cyan} #1}}

\maketitle

\IEEEpeerreviewmaketitle

\input{chapter/ch0_abstract}
\input{chapter/ch1_introduction}

\input{chapter/ch2_definitions}
\input{chapter/ch3_process}
\input{chapter/ch4_layer}

\input{chapter/ch5_simulation}
\input{chapter/ch6_conclusion}
\input{chapter/acknowledgment}

\bibliographystyle{IEEEtran}
\bibliography{bibliography}

\end{document}

%% file: chapter/ch0_abstract.tex
\begin{abstract}
    Trajectory data analysis is an essential component for highly automated driving. Complex models developed with these data predict other road users' movement and behavior patterns. Based on these predictions --- and additional contextual information such as the course of the road, (traffic) rules, and interaction with other road users --- the highly automated vehicle (HAV) must be able to reliably and safely perform the task assigned to it, e.g., moving from point A to B. Ideally, the HAV moves safely through its environment, just as we would expect a human driver to do. However, if unusual trajectories occur, so-called trajectory corner cases, a human driver can usually cope well, but an HAV can quickly get into trouble. In the definition of trajectory corner cases, which we provide in this work, we will consider the relevance of unusual trajectories with respect to the task at hand. Based on this, we will also present a taxonomy of different trajectory corner cases. The categorization of corner cases into the taxonomy will be shown with examples and is done by cause and required data sources. To illustrate the complexity between the machine learning (ML) model and the corner case cause, we present a general processing chain underlying the taxonomy.
\end{abstract}

%% file: chapter/ch1_introduction.tex
\section{Introduction}
    A vehicle equipped with sensor-based assistance systems is driving toward a sharp corner. Suddenly an oncoming vehicle with increased speed appears and cuts the corner. The car driver reduced his speed as a precaution because this dangerous curve was well known to him, and the driver was able to brake just in time to prevent worse. Situations like this, or situations where road users disregard the right of way, are known to every driver. Therefore HAVs must be able to predict trajectories with high reliability. This includes assessing the behavior of other road users and the interaction between them to achieve action planning that can handle such critical situations in the best possible way. Corner cases are usually extremely critical but (fortunately) very rare situations, which makes them very important for training and validating ML methods. As described in the example, many drivers experience these critical situations, but it is hard to fully define them due to their high variance in appearance. The authors in \cite{Heidecker2021b} and \cite{Breitenstein2020} found definitions mainly for the perception tasks, but the domain of action, reaction, and interaction in the context of trajectories in space and time is only marginally addressed by these works. This article examines this domain and specifically trajectories in more detail. The main contributions are:
    \begin{itemize}
        \item Systematization and definition of trajectory corner cases,
        \item Breakdown for corner case situations for trajectory datasets resulting in 31 categories,
        \item Application of proposed systematization on selected use-cases for filtering of corner case and their targeted synthetic generation.
    \end{itemize}
    In Sec.~\ref{sec:definitions}, we introduce different terms regarding trajectories and corner cases, from which we derive our definition of trajectory corner cases. Sec.~\ref{sec:pipeline} contains a process pipeline that may be used to classify corner cases for both human road users and HAVs, considering examples related to trajectories. Using the taxonomy explained in Sec.~\ref{sec:taxonomy} this work provides a high-level approach to gain a better understanding of how a system can deal with driving hazards. Sec.~\ref{sec:application} shows the application of the taxonomy on simulated and real data.

%% file: chapter/ch2_definitions.tex
\section{Definitions}\label{sec:definitions}
    Before discussing the definition of corner cases in trajectories, we present the probabilistic and generative trajectory model and how corner cases are defined in the literature. We then merge these two substantive points, trajectory model and corner cases, and present our definition of trajectory corner cases.

    \subsection{Generative Probabilistic Trajectory Model} \label{subsec:trajectory_model}
        \begin{figure}
            \centering
            \includegraphics[width=0.6\columnwidth]{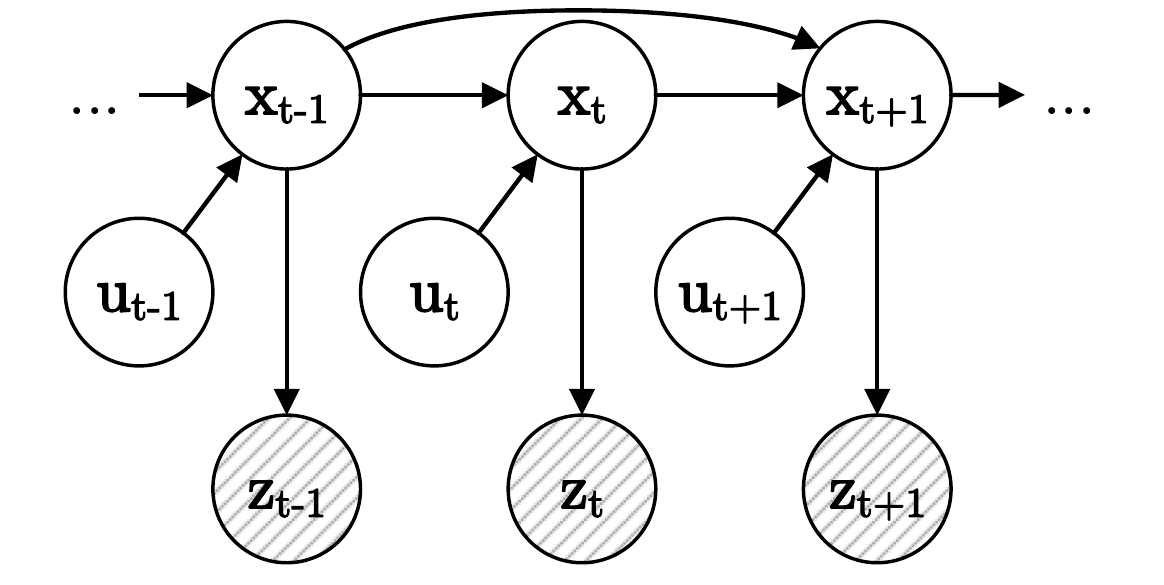}
            \caption{Probabilistic trajectory model characterized as dynamic Bayes network~\cite{TBF05}. The variable $\bm{z}_t$ describes the observable, noisy version of the underlying latent state $\bm{x}_t$, which is influenced by a control-input $\bm{u}_t$.}
            \label{fig:trajectory_model}
        \vspace*{-.5cm}
        \end{figure}
        In the following, we introduce a probabilistic trajectory model which is the basis of our corner case definition. Environments and technical systems are often characterized in terms of states $\bm{x}$~\cite{TBF05}. A trajectory is a time series of states in some metric space which additionally possesses an inherent spatial smoothness that directly links two neighboring states $\bm{x}_{t-1}$ and $\bm{x}_t$ in time~\cite{Bie21}. Here, the subscript $t$ denotes the time dependency. A trajectory displays the dynamics of a system under observation in such a state space. Formally, we define a trajectory of length $N \in \mathbb{N}$ in a state space as the set $\bm{X} = \left( \bm{x}_t\ |\ t \in [1, \ldots , N] \right)$. Without loss of generality, we consider trajectories in state spaces comprising 3D-poses $\bm{x}_t \in  SE(3)$ in the special Euclidean group consisting of position and the orientation of the system under observation. Note that in many applications in highly automated driving the state space often also comprise single- or multi-order derivatives, e.g., the velocity and acceleration.
        We model the dynamics of the system under observation probabilistically defined as  $p\left(\bm{x}_{t+1}\ |\ \bm{x}_t, \ldots , \bm{x}_1, \bm{u}_t, \mathcal{C}_d \right)$.
        The evolution of the state $\bm{x}_{t+1}$ may depend on former states $\bm{x}_t, \ldots , \bm{x}_1$ as well as optional control-inputs $\bm{u}_t$, e.g., acceleration, braking, or steering input variables. Moreover, $\mathcal{C}_d$ denotes additional context, e.g., road- and weather-conditions, (traffic) rules, age of driver, or the intentions of the driver and the desired target. In general, we do not have access to the true model underlying the actual trajectory nor do we have knowledge about all context variables. Instead, in practice, we approximate the model as well as the influencing parameters directly in terms of a parametric model or indirectly in terms of a dataset. Furthermore, in most case we cannot observe the true state $\bm{x}_t$ directly, but instead often measure a noise version $\bm{z}_t$ of the actual underlying state. Without loss of generality, we assume a simple probabilistic measurement model $p(\bm{z}_t\ | \bm{x}_t, \mathcal{C}_m)$,
        where $\mathcal{C}_m$ denotes the context relevant for the measurement model. The latter could for example be a potential obstruction. A graphical model of our generative probabilistic trajectory model is depicted in Fig.~\ref{fig:trajectory_model}. As before, we do not have access to the exact model governing the measurement generation process. We define an observed trajectory $\bm{Z} = \left( \bm{z}_t\ |\ t \in [1, \ldots , N] \right)$. A trajectory dataset $\mathcal{D}$ consists of multiple observed trajectories. Note that we do not know the true trajectory model nor the actual context involved. However, we can see observed trajectories as samples from this probabilistic trajectory model.

    \subsection{Corner Case Definition}\label{sec:ccterms}
    In the following, we first review different approaches and perspectives to define the term \textit{corner case} before introducing our definition for trajectory corner cases.
    \subsubsection{Corner Case Definitions in Literature}
        Koopman~et~al.~\cite{Koopman2019} look at edge and corner cases from the ML testing and validation perspective. Both edge and corner cases are rare cases, e.g., driving or traffic situations, that will occur only occasionally, wherein \textit{rare} means that these cases often occur at large-scale deployed systems, e.g., full-scale deployed vehicle fleet. An edge case is a situation that requires specific attention during the design time of the system at hand; e.g., these may be sensor values that are out of the predefined bounds or missing values. In contrast, a corner case results from the combination of several \enquote{normal} input parameters or situations that coincide simultaneously, thus representing a rare or never considered case or scene \cite{Koopman2019}.
        This definition is the basis of many test frameworks, such as DeepXplore~\cite{Pei2017} or DeepRoad~\cite{Zhang2018}. These methods use differential testing to generate synthetic test cases systematically, also referred to as corner cases. They define corner cases as those in which neural network predictions are indecisive. Hence, the network predicts \enquote{I don't know}~\cite{Platon2020}. %This to the unpredictability of corner cases.
        Houben~et~al.~ \cite{Houben2021} give a general definition of corner cases stating that \enquote{inputs that result in unexpected or incorrect behaviour of the AI [artificial intelligence] function are defined as corner cases.} This defines corner cases as situations where the ML model is unable to understand the situation correctly. Others consider mere erroneous or incorrect behavior as corner cases~\cite{ZHM+22}. From the perspective of model training, especially active learning~\cite{kottke2021optimal}, corner cases are the instances, e.g., trajectories that yield the greatest possible learning progress. Following this definition, an instance identified as a corner case and subsequently used for model (re-)training is no longer a corner case. According to Bolte~et~al.~a corner case in the context of automated driving and computer vision is given \enquote{if there is a relevant object (class) in a relevant location that a modern automated vehicle cannot predict}~\cite{Bol19}.

    \subsubsection{Corner Case in Trajectory Datasets}
        Assume we are given a trajectory dataset $\mathcal{D}$, e.g., of pedestrian trajectories. Let us imagine, we are given a new trajectory $\bm{Z}_{\text{ped}}$ of a pedestrian intending to cross a street and a vehicle $\bm{Z}_{\text{veh}}$ approaches from far away. Based on the dataset $\mathcal{D}$ (in the case of humans, these are years of life experience) and other available contextual information, e.g., the environmental circumstances such as zebra crossings and (traffic) rules to be obeyed, we as humans have an expectation of how a pedestrian trajectory should look like. In other words, we have an estimate of the pedestrian's dynamics and the measurement model (cf.\ Sec.~\ref{subsec:trajectory_model}).
        However, if a pedestrian stops in the middle of the road to tie her shoes, the pedestrian's trajectory deviates from the expected trajectory of a pedestrian crossing the road. This suggests that our estimated trajectory model is not explaining the observed trajectory correctly, as something anomalous and unpredictable happened.
        To what extent a trajectory is unexpected depends strongly on the existing prior knowledge and available context, e.g., in our dataset $\mathcal{D}$, we have seen kneeling pedestrians frequently, but not yet in combination with the context of kneeling in the middle of the road. Two common circumstances might lead to an extremely unusual trajectory (since rare). If our task is to train an ML model to predict a pedestrian trajectory as good as possible~\cite{Bie21}, then such a trajectory can be considered highly relevant and thus a corner case. However, this is rather irrelevant from the perspective of save automated driving as the HAV is still far away, and there is no interaction between the two. Depending on our task not every anomalous trajectory $\bm{Z}_\text{ped}$ has to be a corner case. Imagine the vehicle is close by, then due to the interaction of the two road users (given already by the mere presence of the two), such a scene and thus the trajectories of the involved road users experience a completely new assessment. A correct model of the pedestrian's (future) trajectory is now also highly relevant for avoiding an accident. Another example of such a corner case is a vehicle that passes a red traffic light. The plain trajectory of such a vehicle will not differ from a rule-compliantly driving vehicle. It becomes a corner case only if the context (here the red traffic light) and the disregard of the traffic rules are taken into account. To capture such a corner case, additional information is needed. We conclude that to assess whether a trajectory is a corner case, we must consider the interaction with other objects and agents, i.e., road users, (traffic) rules, and its relevance for the task at hand. Likewise, noisy trajectories with outliers or missing values do not necessarily have to be a corner case. Besides studying relevance, missed measurements, outliers, and missing values can be modeled by an appropriate probabilistic measurement model so that they do not necessarily have to be unexpected nor abnormal.
        Concluding this into the definition of trajectory corner case within a trajectory dataset $\mathcal{D}$ as follows: \\

        \textbf{Definition Corner Case Trajectory:}
        \textit{A corner case in a trajectory dataset $\mathcal{D}$ is a highly relevant but mostly very rare and anomalous trajectory.
        The relevance of a trajectory is mainly determined by the interaction with other agents (e.g., road users), the surrounding environment, norms and (traffic) rules, and most importantly, by the task at hand.}

    \subsection{Anomalous Trajectories}
        The terms \textit{outlier}, \textit{anomaly}, and \textit{novelty} are often used interchangeably and are all strongly related to corner cases~\cite{Heidecker2021b}. In the case of trajectories, we speak of an outlier when one or more measurements deviate so strongly that we assume they were generated by a different underlying process~\cite{Hawkins1980}. An anomaly is something that deviates from the standard, average, or expected, whereas an anomalous trajectory refers to the trajectory with local or global differences from most other average trajectories when measured by some similarity metrics~\cite{GLG+22}. Hence, by definition, an anomalous trajectory is rarely represented within the trajectory dataset at hand. An agglomeration of anomalies as clusters is called novelty~\cite{GRUHL2021}. The challenge of detecting anomalous trajectories is referred to as anomalous trajectory detection~\cite{GLG+22}. Among others, we can distinguish between the following approaches to detect anomalous trajectories: clustering-~\cite{Eldawy2020}, distance-, density-based~\cite{Heidecker2019}, and deep-learning-based~\cite{liu2020online} anomalous trajectory detection.

    \subsection{Interaction}
        Markkula~et~al.~\cite{markkula2020} defines \textit{interaction} as a situation in road traffic in which the behavior of at least two road users can be interpreted in such a way that both may occupy the same spatial region at the same time. The interaction of road users is the precondition for traffic conflicts, e.g., situations that can potentially lead to an accident~\cite{harris1968traffic} and hence a basic building block for determining the relevancy of trajectories. Moreover, modeling the interaction of road users is essential to understand the behavior, e.g., their trajectory, of individual road users in traffic. An example of these are the Gipps or intelligent driver model~\cite{TKT12}, which describes the \enquote{following} behavior of individual vehicles in traffic. There are also numerous interaction models for persons, e.g., the social force model for modeling pedestrian dynamics~\cite{helbing1995socialforcemodel}. Recently, data-driven approaches also explicitly learn interaction patterns to improve prediction, e.g., the social long short-term memory approach~\cite{alahi2016sociallstm}. Understanding and modeling the interaction with the environment and between the individual road users is crucial to judge whether a trajectory is anomalous, i.e., contains unexpected interaction patterns, and relevant, i.e., due to a potential conflict, hence, a corner case.

    \subsection{Relevancy} \label{subsec:relevancy}
        Following our corner case definition, we might be tempted to ask what is relevant and how we can measure the relevance to a given trajectory dataset. Whether a trajectory is relevant strongly depends on the task at hand. Therefore, relevance must be defined on a task-specific basis, making a general definition difficult. In this section, we review different perspectives in the realm of highly automated driving regarding how relevance is defined and measured.

    \subsubsection{Traffic-Safety Perspective}
        Intuition tells us that the relevancy is related to the hazard or risk of a perceived trajectory. Hence, to assess the relevancy of a trajectory and thus identify corner cases, we must inspect the spatial and temporal context of objects in the vicinity of automated vehicles. In other words, the relevancy of trajectories in the highly automated driving domain is correlated with the chance of a crash or near-crash. Therefore, we must also take the interaction of trajectories into account. From the perspective of traffic safety, a trajectory is relevant if it is critical, i.e., the trajectory or multiple trajectories need explicit consideration of potential (crash) mitigation strategies~\cite{ZTT+21}.
        Many other measures aim to quantify the relevancy of trajectories in terms of criticality or proxy measures such as maximum relative speed, time-to-X, e.g., time-to-break, time-to-collision, time-to-touch~\cite{HKS05}, the distance of closest encounter~\cite{Egg14}, time-to-closest encounter~\cite{Egg14}, and threat-assessment~\cite{DCO+19}. Kamran~et~al. define relevant objects in automated driving, which can be transferred to trajectory datasets. According to them, relevant objects are defined by their possible conflict zone with an (automated) ego vehicle. The relevance of trajectories is directly derived from the possible conflict between them~\cite{kamran2020risk}. The authors in \cite{GLG+22} define a metric to measure the relevance by combining an interaction and anomaly score. However, again which measure is best suited for the detection and quantification of relevant trajectories depends on the precise objective.

    \subsubsection{Machine-Learning Perspective}
        The prediction is the base of the environment abstraction which is used for decision-making and trajectory planning. The entire processing pipeline consists of many smaller and bigger tasks while every one of them holds their own relevancy. The relevance of individual sensor data, e.g., trajectories and all knowledge abstracted from them are already determined at this point since the selected methodology, model architecture, parameters and hyperparameters for each task have been trained or adjusted accordingly. In self-learning systems such as active learning, the situation is somewhat different. The chosen hyperparameters also determine whether the model classifies the trajectory as relevant for further training or not. Whereas the choice of methodology --- choosing between autoencoder and GAN (generative adversarial network)~\cite{Goodfellow2016} or deciding between a simple detector and a classifier --- essentially depends on the task.
        However, at the end the implemented model architecture determines how the data is processed. For example, the choice of the autoencoder latency directly impacts which information is retained or irretrievably lost. Nevertheless, other design decisions like convolutional layer, fully connected layer, or pooling layer also influence which data is considered relevant. This behavior is also reinforced by the fact that specific feature are weighted more strongly by optimization or model training. For example, if we look at the loss function, which has a direct impact on the model weights during training, the mean squared error loss ensures that outliers are taken into account to a much greater extent than by the Huber loss, which is less sensitive to outliers.

    \subsubsection{Rule-Compliance}
        The goal of rules is to ensure that processes always follow the same pattern and that no participant is put in danger. They provide predictability about the behavior of others. This is particularly essential in road traffic where the violation of traffic rules may result in a dangerous situation. Hence, rule violation within a trajectory is directly coupled with the relevancy of that trajectory; for example, a violation of the right of way may result in dangerous situations. Even the fine catalog offers a possibility to quantify the relevancy of non-rule-compliant trajectories. However, there are indeed situations where an automated vehicle has to perform a \enquote{controlled}-rule exception. Imagine an automated vehicle on the street with a solid center stripe waiting behind a parked ambulance due to a rescue mission. In this case, the following vehicles need to enter the oncoming lane, even if the road marking forbids that. This is an easy task for a human driver but a challenging and relevant corner case for the development of HAVs.

%% file: chapter/ch3_process.tex
\section{Process Pipeline for Road Users} \label{sec:pipeline}
\begin{figure*}
  \centering
  \includesvg[inkscapelatex=false, width=\textwidth]{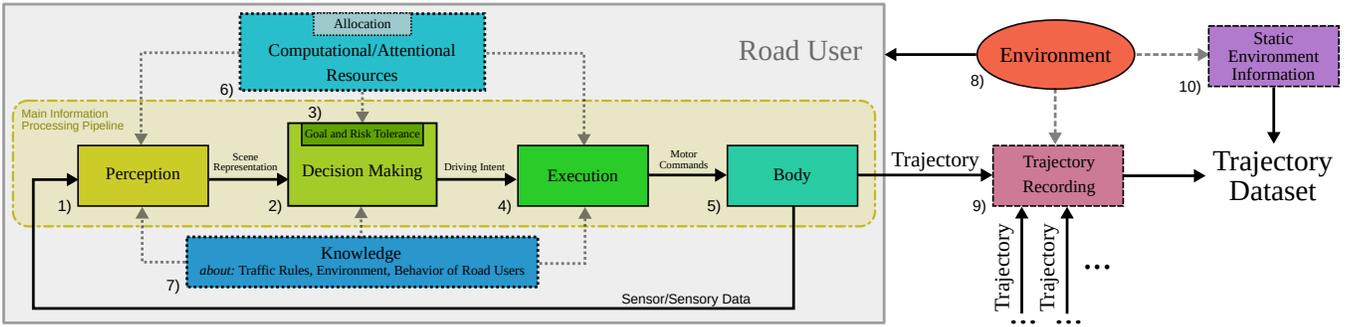}
  \caption{Schematic of the processing pipeline for both a human road user or an automated vehicle}
  \label{fig:process_pipeline}
  %\vspace{-1em}
\vspace*{-.5cm}
\end{figure*}

A trajectory corner case taxonomy is helpful to understand critical situations better and consider the reasons leading to the incident.
If the taxonomy is used to analyze the misbehavior of an HAV in a specific situation, the first question that arises will be: which part of the system malfunctioned? If on the other hand, the performance of such a system has to be evaluated more generally in rare and unusual traffic situations, the actual behavior of the other road users is the center of attention. But even then, it is obvious that errors in different parts of the human or machine driver's information processing system will result in completely different behaviors and trajectories.
Therefore, it makes intuitive sense to categorize corner cases by the origin in the driver's information processing system, whether this is the human brain or the software of an HAV.
A unified model of the information processing pipeline for automated vehicles, human drivers, and VRUs (e.g. pedestrians, motorcyclists, and cyclists) leads to a more comprehensive taxonomy while at the same time allowing analysis or classification independent of the level of automation of the road user under consideration.

The task of highly automated driving is typically accomplished by dividing it into sub-problems and corresponding software modules. On the coarsest scale, we can distinguish between perception, decision making, trajectory planning, and control. The \textit{Main Information Processing Pipeline} in Fig.~\ref{fig:process_pipeline} mirrors this partition. Trajectory planning and control are here combined in the \textit{Execution} module. Published system architectures are compatible with this processing pipeline model, cf.~\cite{tacs2017making,9046805,Berger2014,MatthaeiMaurer2015}, although the names of the modules vary.

On this high level of abstraction, the stages of information processing in humans bear many similarities to the ones in highly automated driving software. We use the model of human information processing stages by Wickens~\cite{wickens2016} as reference for human road users. This very general model is widely used and generally accepted in traffic psychology and traffic safety research~\cite{Shinar2017}. We picked this model over other well-established models from traffic psychology and human factors as many models presume specific cognitive mechanisms that are not applicable to HAVs~\cite{endsley1995,anderson2004integrated,psi} (e.g. schema activation) in addition to the core components described by Wickens.

The \textit{Main Information Processing Pipeline} (highlighted in Fig.~\ref{fig:process_pipeline}) consists of five components:
\begin{enumerate}
\item \textit{Perception}: Processes sensor or sensory organ data to generate an (explicit or implicit) scene representation. This includes the recognition and interpretation of road boundaries and street signs as well as the detection and behavior prediction of other road users.
\item \textit{Decision Making}: Given the scene representation, this component decides which actions to take. The outputs are not motor commands, but high-level driving intents, e.g. whether to stop or not, whether to overtake, to yield, which road or lane to follow and at which speed.
\item \textit{Goal and Risk Tolerance}: The decisions depend on the goals
of the human or machine driver and the level of risk they are willing to take to achieve these goals.
\item \textit{Execution}: The translation of the driving intent to motor commands. For an HAV, this encompasses local trajectory optimization and control.
\item \textit{Body}: The physical entity of the road user, e.g. the body of a pedestrian or vehicle. The body also includes the actuation, i.e. the muscles or motors that translate motor commands to movements.
\end{enumerate}
Two additional components that act on all main components complete the road user model:
\begin{enumerate}[resume]
\item \textit{Computational/Attentional Resources}: The information processing capacity of both human brains as well as computers is limited. In psychology, these resources are termed attention resources (or simply attention)~\cite{wickens2016}. These resources can be redistributed within certain limits in both humans and computers to adapt to rapidly changing demands. In humans, this happens through a mechanism which is also just termed attention~\cite{wickens2016,wickensappliedattention}, and in HAV through scheduling and attention mechanisms in deep neural networks.
\item \textit{Knowledge}: All information processing stages require (a priori) knowledge about traffic rules, the behavior of other road users, and the physical environment (e.g. map data for localization or navigation). This knowledge can be explicitly stored in databases or long-term memory; or it can be implicit in algorithms, parameters, ML models and the datasets they are trained with, and (human) learned behaviors and experiences.
\end{enumerate}
Even though more than 90\% of all traffic accidents are caused by human error~\cite{Sabey1980}, the behavior of the road users still cannot be analyzed independently of their \textit{Environment}. The road users sense the \textit{Environment} through their sensors or sensory organs, but the \textit{Environment} can also impair the observability of other road users through occlusions or adverse weather conditions. Road conditions can fundamentally change the relationship between the motor inputs and the resulting road user trajectories. Road markings, street signs, and obstructions determine which trajectories are legal and which trajectories are possible.

%% file: chapter/ch4_layer.tex
\section{Taxonomy for Corner Cases in Trajectory Data for Automated Driving}\label{sec:taxonomy}

\begin{table*}[htbp]
    \centering
    \begin{tabulary}{\textwidth}{|>{\bfseries}L|>{\em}C|>{\em}C|>{\em}C|>{\em}C|}
        \hline
        \normalfont{\textbf{stage}} & \normalfont{\textbf{Ego Trajectory}} & \normalfont{\textbf{Ego Trajectory \& Other Road Users}} & \normalfont{\textbf{Ego Trajectory \& Environment}} & \normalfont{\textbf{Ego Trajectory \& Other Road Users \& Environment}} \\
        \hline\hline
        \rowcolor{perception} 1)~Perception & Emergency break, because deceleration of the car ahead was not registered & Near collision because of overlooked crossing traffic & Driving wrong direction in one-way street because of overlooked sign & Taken right of way because of overlooked sign\\
        \hline
        \rowcolor{decision} 2)~Decision Making  & High velocity in sharp corner & Near collision in merge scenario due to time pressure & Planned disregard for a puddle that covers a large hole & Failure to understand precedence at a complex crossing \\
        \hline
        \rowcolor{goal} 3)~Goal and Risk Tolerance & Kick-down start & Tail gating & Cutting corners & Willingly taken right of way \\
        \hline
        \rowcolor{execution} 4)~Execution & Stall engine & Near rear-end collision because of insufficient braking & Scrape obstacle when parking & Rolling too far into an intersection due to insufficient braking while yielding\\
        \hline
        \rowcolor{body} 5)~Body & Tire burst & Near rear-end collision due to brake failure & Near collision due to failing headlights & Taken right of way due to bad visibility of other vehicle because of broken windshield or camera optics \\
        \hline\hline
        \rowcolor{attention} 6)~Computational and Attentional Resources &  \multicolumn{4}{c|}{Any of the examples from stages 1 -- 5} \\
        \hline
        \rowcolor{knowledge} 7)~Knowledge & High velocity in surprisingly sharp corner &  Pedestrian at crosswalk looking away is interpreted as yielding  & U-turn in no passing area
 & Slow approach of crossing interpreted as giving right of way \\
         \hline\hline
        \rowcolor{environment} 8)~Environment & Crash into blown over tree & Near collision due skidding on ice & Leaving the road due to Aquaplaning & Taken right of way due to view obstruction \\
        \hline\hline
         \rowcolor{recording} 9)~Trajectory Recording & \multicolumn{4}{c|}{\textit{Noisy IMU measurements or failed tracking of ego road user in camera image}}\\
        \hline
        \rowcolor{environmentinfo} 10)~Static Environment Information & \multicolumn{2}{c|}{\cellcolor{lightgray}{}}   & \multicolumn{2}{c|}{\textit{Missing priority sign in dataset}}  \\
        \hline
    \end{tabulary}
    \caption{Trajectory corner case taxonomy with example situations. The rows describe the system component in which the corner case originates. The columns identify the trajectory corner case class describing the kind of data required to detect the corner case. Ego refers to the road user that causes the corner case (multiple ego vehicles are possible if multiple trajectories are anomalous, resulting in multiple assigned categories for one corner case situation). The colors and stage number correspond to those of the components in the processing pipeline in Fig.~\ref{fig:process_pipeline}.}
    \label{tab:traj_categorization}
\vspace*{-.5cm}
\end{table*}

Based on the process pipeline defined in Fig.~\ref{fig:process_pipeline}, a taxonomy is developed that categorizes trajectory corner cases according to two criteria: what data is required to detect the corner case and which system component caused the corner case. Together, these two criteria span Tab.~\ref{tab:traj_categorization} resulting in 31 categories. The columns correspond to the four classes which describe the data required for recognition and the rows where in the system the corner case originated.

\subsection{Columns}
If we want to detect trajectory corner cases, we have to first consider what data is available and also what kind of data is required to classify specific kinds of corner cases we might be interested in. This might also restrict which datasets are suitable for training and has implications on the hard- and software required to detect corner cases in the wild.

Some corner cases are detectable from just the trajectory of the road user who is responsible for the corner case (subsequently termed ego). These corner cases fall into the first class \textit{Ego Trajectory}. Corner cases of this class are typically characterized by unusual speed or acceleration values. An example of this is a car that drives at high velocity in a sharp corner, resulting in an unusually high lateral acceleration. Collisions and near-collisions with violent evasive maneuvers are also detectable by the unusual acceleration values alone and therefore also fall into this category. Inertial sensors are often sufficient to detect corner cases of this class.

Corner cases that result from the interaction of multiple road users, but do not result in a collision or an evasive maneuver, require more than one trajectory for classification. These corner cases fall into the second column of our taxonomy.

Many trajectory corner cases however can only be identified if additional environmental context is available. These corner cases fall into the third and fourth columns. Environmental context refers to any kind of information beyond the trajectory data, e.g. lane topology, obstacles, and traffic rules per lane (e.g. speed limit). Most traffic rule violations are only detectable with proper environmental context. If the ego trajectory and the environmental context are enough to detect the corner case, it is assigned to the third column. If both environmental context and the trajectories of other road users are required to detect the corner case, it is assigned to the fourth class. This is the case for violations of more complex traffic regulations like priority rules.

\subsection{Rows}
The rows of our taxonomy specify which system component of the ego-vehicle failed or is the cause of the unpredictable behavior, where each row in Tab.~\ref{tab:traj_categorization} corresponds to one of the system components described in Sec.~\ref{sec:pipeline}. This aspect of our taxonomy is similar to some previous taxonomies from traffic safety research~\cite{treat1977tri,STANTON2009227, KHATTAK2021105873}. These previous taxonomies are, however, strictly human driver error taxonomies as they include mechanisms or information processing stages that have no equivalent in HAVs which is comparable in terms of the behavior in case of a fault (e.g. \enquote{memory and recall}~\cite{STANTON2009227}, \enquote{fatigue}~\cite{treat1977tri}, \enquote{experience}~\cite{KHATTAK2021105873}).
Depending on the use case, a further breakdown into subcategories can make sense. Beneficial is the distinction between \textit{Computational/Attentional Resources} and their allocation and between different types of \textit{Knowledge}, e.g., a priori knowledge about the behavior of other road users, the environment, and traffic rules.

It is in general not possible to unequivocally assign a single category to all situations, as multiple anomalies with distinct origins in the information processing pipelines can appear at different points of the trajectory. These anomalies often are also dependent on each other, further complicating a consistent categorization: If a driver takes another car's right of way because they are blinded by the low winter sun, is this an \textit{Environment} corner case due to the unfavorable lighting conditions? Or a \textit{Perception} corner case, as a more observant driver might have been able to see the other vehicle in time? Or even a \textit{Goal and Risk Tolerance} corner case, as a more careful driver might have approached the intersection more carefully given the environmental conditions?
This is, however, not an issue specific to our corner case taxonomy. This problem was first pointed out by Senders and Moray in the context of human error taxonomies~\cite{senders1991human}. In analogy, it also applies to human and HAV corner case taxonomies: Causal chains make selecting a single cause of the corner case context-dependent, i.e., on the application or the user's judgment. It, therefore, can be preferable to not assign a single, but multiple categories to each corner case situation.

Another issue of attributability arises if the taxonomy is used for the classification of corner cases in trajectory datasets. The \textit{Computational/Attentional Resources} and \textit{Knowledge} components are frequent error sources and therefore important categories if a specific road user system is analyzed, but are not useful for the classification of corner cases in black box systems. Errors in these two components can manifest in any of the \textit{Main Information Processing Pipeline} components. It is, therefore, often impossible to determine if the \textit{Computational/Attentional Resources} or \textit{Knowledge} components are the error source if only the trajectory (and environmental context) are available. Instead, for classification tasks errors in these categories have to be assigned to the components in which they manifest.

On the other hand, two additional categories become relevant for dataset classification tasks: Firstly, measurement errors like excessive noise might corrupt the trajectories during \textit{Trajectory Recording}; secondly, mistakes could creep into the \textit{Static Environment Information} or metadata the dataset might be enriched with. In these cases, the trajectories in the dataset could be considered corner case trajectories, although the actual ground truth trajectories might not represent corner cases at all.

%% file: chapter/ch5_simulation.tex
\begin{figure*}[htp]
    \centering
    \begin{subfigure}{0.32\textwidth}
        \includegraphics[width=1\linewidth, trim={0 33px 0 33px}, clip]{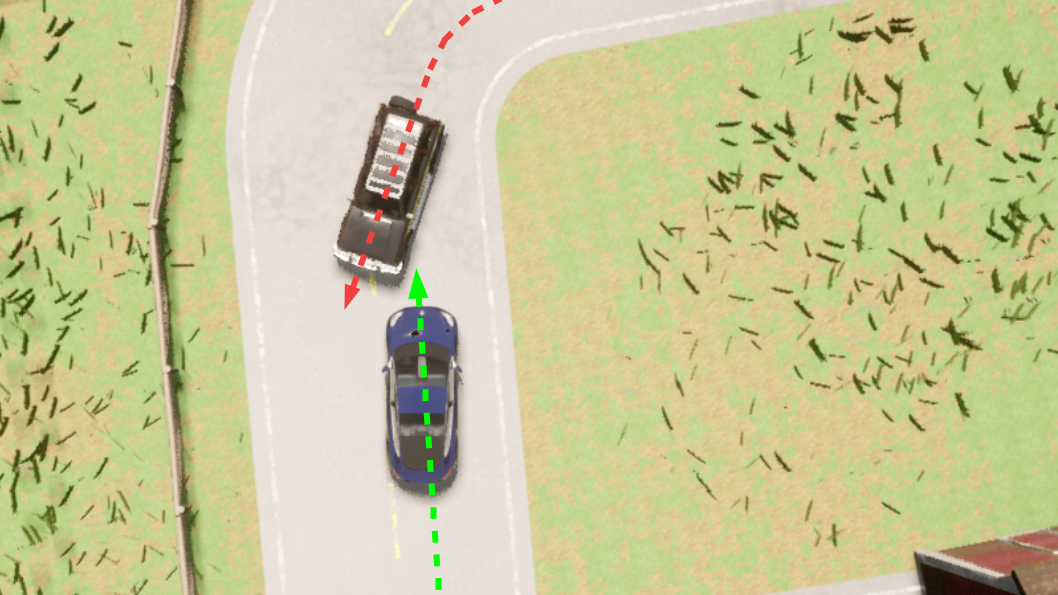}
        \caption{Cutting corners.}
        \label{fig:cutting_corner}
    \end{subfigure}
    \hfill
    \begin{subfigure}{0.32\textwidth}
        \includegraphics[width=1\linewidth, trim={0 33px 0 33px}, clip]{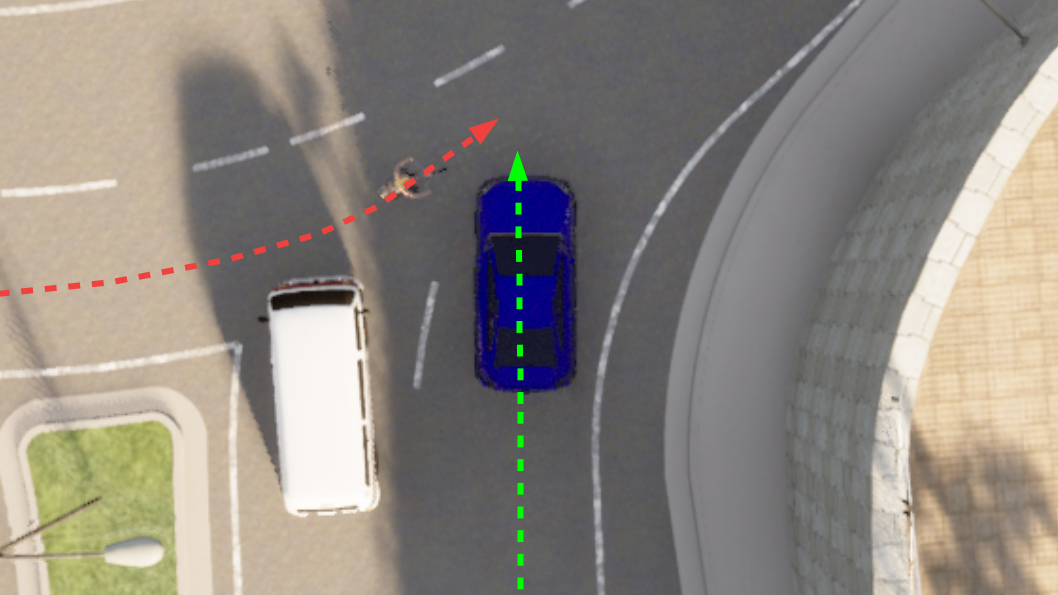}
        \caption{Covering and increased speed.}
        \label{fig:traffic_circle}
    \end{subfigure}
    \hfill
    \begin{subfigure}{0.32\textwidth}
        \includegraphics[width=1\linewidth,trim={0 3px 0 3px}, clip]{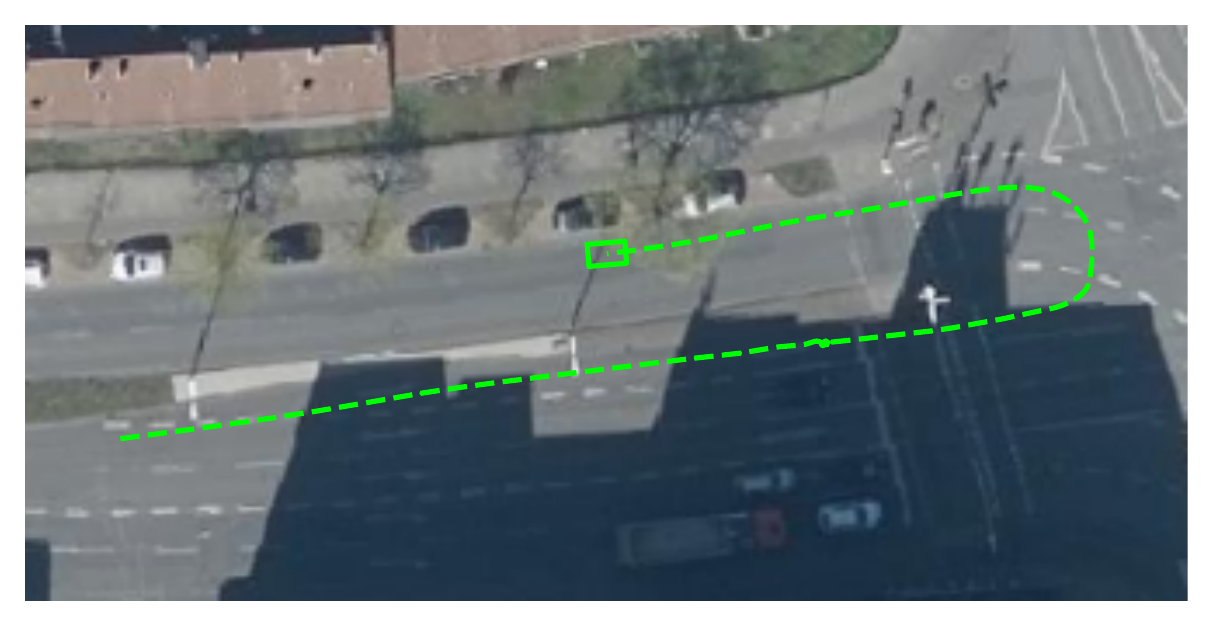}
         \caption{U-turn situation.}
         \label{fig:u-turn}
    \end{subfigure}
    \caption{Applications of the taxonomy on simulated and real world data. (a) Shows an accident event due to crossing the oncoming traffic which is classified as goal and risk tolerance with environmental reference. (b) Accident event between ego vehicle and cyclist caused by a third road user obstructing the view and the increased speed of the ego vehicle. (c) U-turn trajectory from the recordings of the KI Data Tooling project. At the last data point of the trajectory, the bounding box of the object is shown. The background image is an excerpt from the geodata of the State Office for Geoinformation and Land Surveying Lower Saxony (www.lgln.de), \copyright2022.}
    \label{fig:figures}
    \vspace*{-.5cm}
\end{figure*}

\section{Application on Data}\label{sec:application}
    After having defined the taxonomy, this chapter shows how to record and classify corner cases in real or simulation scenarios. The taxonomy is applied on several scenes with varying setups to demonstrate the generalization of the method.

\subsection{Application on Simulation Data}
    This section is about the generation of corner cases in the autonomous driving simulation software CARLA~\cite{dosovitskiy2017carla}. The software is an open source software for data generation and testing of ML algorithms and provides various sensor data to describe the simulated scenario, such as camera, LiDAR, and RADAR. Moreover, the software also delivers ground truth data that is cost-intensive and time-consuming to generate with real data. In addition, it offers the possibility of creating safety-critical situations in the virtual world so as not to endanger road users. In \cite{kowol2022aeye}, a human-in-the-loop approach for generating safety-critical synthetic corner cases is presented. For this purpose, the authors have set up a test rig with two control units so that the same vehicle can be controlled by two human drivers in CARLA. Here, a semantic segmentation network is integrated into CARLA so that one of the drivers (semantic driver) only sees the networks output and thus has to control the vehicle in the CARLA world. The other driver (safety driver) monitors the situation and is supposed to intervene as soon as the semantic driver shows dangerous driving behavior. Interventions by the safety driver indicate poor situational awareness by the segmentation network and provide a safety-critical corner case. During the experiments in \cite{kowol2022aeye}, several corner cases were recorded from which the trajectory data could subsequently be obtained. Two of them are now classified based on the categories defined in Sec.~\ref{sec:taxonomy}.

    In the first scene, an accident occurs even though the ego vehicle is obeying the traffic rules due to the fact that the road user in the opposite lane is driving too fast around the curve and crosses the lane of the ego vehicle, see Fig.~\ref{fig:cutting_corner}. Due to the increased speed, the road user crosses the oncoming lane, so this situation would be classified as \textit{Goal and Risk Tolerance} related to the environmental trajectory.

    In the second scene, a vehicle is driving towards a two-lane traffic circle. Next to this vehicle is a bus, which somewhat obscures the view of the road. While the bus driver allows a cyclist to pass, the vehicle owner collides with him due to impaired visibility, see Fig.~\ref{fig:traffic_circle}. This situation is \textit{not} classified as a \textit{Perception} error, because the cyclist was not perceivable behind the bus. Instead, as the \textit{Decision Making} should account for the possibility of occluded road users with priority, the error is attributed to this category.% This is an \textit{Ego Trajectory \& Other Road Users} corner case, as it is detectable without environmental context due to the intersecting trajectories.

\subsection{Application on Real Infrastructural Data}
    This section is about the detection of corner cases in real trajectory data. The investigated trajectory data were extracted from the recordings of a 14 video camera setup from the Application Platform for Intelligent Mobility (AIM) Research Intersection of the German Aerospace Center, located in Braunschweig~\cite{KnakeLanghorst2016}. The dataset was collected as part of the project KI Data Tooling~\cite{kidt} and consists of recordings with a total length of 13 hours. As for the simulation data, multiple corner cases are included in the dataset, and one of them is exemplary described in the following.

    The corner case \textit{U-turn in no-passing area} is part of the stage \textit{Knowledge} in the class \textit{Ego Trajectory and Environment}. Although a no-turning sign was installed in the west of the signalized research intersection, road users keep executing U-turns there. The KI Data Tooling dataset contains several U-turns, from which one of them is visualized in Fig.~\ref{fig:u-turn}.

    The U-turn trajectory was identified by employing virtual loops as in~\cite{Junghans.2021}. Therefore the loops were placed manually at the entry and exit of the junction. Afterward, all intersections between the trajectories of the dataset and the virtual loops were calculated. Finally, the U-turn trajectory can be detected because, compared to other trajectories, it intersects both virtual loops.

%% file: chapter/ch6_conclusion.tex
\section{Conclusion}
    In this article we proposed a definition of corner case trajectories in the context of automated driving and described its reference to related terms like \textit{anomalous}, \textit{interaction}, and \textit{relevance}. Furthermore, we presented a model of a processing pipeline, which is applicable to a human or automated vehicles and can be employed to categorize the causes of corner cases. Finally, we defined our taxonomy for corner case trajectories by combining the presented model with the different types of data sources that are required to detect the corner case. To demonstrate the use of the taxonomy, we gave examples of corner case trajectories in real and synthetic data.

%% file: chapter/acknowledgment.tex
\section*{Acknowledgment}
    This work results from the project KI Data Tooling (19A20001O) funded by the German Federal Ministry for Economic Affairs and Climate Action (BMWK).
    We gratefully acknowledge the advise of Jonas Imbsweiler.